\documentclass[10pt,twocolumn,letterpaper]{article}

\usepackage{iccv}
\usepackage{times}
\usepackage{epsfig}
\usepackage{graphicx}
\usepackage{amsmath}
\usepackage{amssymb}
\usepackage{booktabs}
\usepackage{multirow}
\usepackage{caption}
\usepackage{subcaption}
\usepackage[table]{xcolor}
\usepackage[scale=0.82]{FiraMono}

\usepackage{enumitem}

\usepackage[utf8]{inputenc} 
\usepackage[T1]{fontenc}    
\usepackage{url}            
\usepackage{booktabs}       
\usepackage{amsfonts}       
\usepackage{nicefrac}       
\usepackage{microtype}      
\usepackage{xcolor}         

\usepackage{color}
\usepackage{amssymb}
\usepackage{pifont}
\usepackage{multirow}
\usepackage{graphicx}
\usepackage{gensymb}

\usepackage{amsmath}
\usepackage{subcaption}
\usepackage{bm}
\usepackage{mathrsfs}
\usepackage{makecell}
\usepackage{colortbl}
\usepackage{wrapfig}
\usepackage{float}
\usepackage{soul}
\usepackage[permil]{overpic}
\usepackage{tabularx}
\usepackage{lipsum}
\usepackage{siunitx}
\usepackage{textcomp}
\usepackage{enumitem}
\usepackage{booktabs}
\usepackage{array}\newcolumntype{C}{>{\centering\arraybackslash}X}

\usepackage{listings}
\usepackage{algorithm}
\usepackage{bbold}

\usepackage[breaklinks=true,bookmarks=false]{hyperref}

\iccvfinalcopy 

\def\iccvPaperID{****} 

\ificcvfinal\fi

\makeatletter
\renewcommand{\paragraph}{%
  \@startsection{paragraph}{4}%
  {\z@}{1.25ex \@plus 1ex \@minus .2ex}{-1em}%
  {\normalfont\normalsize\bfseries}%
}
\makeatother

\newcommand{\tablestyle}[2]{\setlength{\tabcolsep}{#1}\renewcommand{\arraystretch}{#2}\centering\footnotesize}
\newcolumntype{x}[1]{>{\centering\arraybackslash}p{#1pt}}
\newcolumntype{y}[1]{>{\raggedright\arraybackslash}p{#1pt}}
\newcolumntype{z}[1]{>{\raggedleft\arraybackslash}p{#1pt}}
\definecolor{green}{HTML}{39b54a}  
\definecolor{red}{HTML}{ea4335}  
\definecolor{white}{HTML}{ffffff}  
\definecolor{grey}{HTML}{bfbfbf}  
\definecolor{blue}{HTML}{0000FF}  
\newcommand{\hlg}[1]{\cellcolor{green!20}{#1}}

\usepackage{etoolbox}
\makeatletter
\AfterEndEnvironment{algorithm}{\let\@algcomment\relax}
\AtEndEnvironment{algorithm}{\kern2pt\hrule\relax\vskip3pt\@algcomment}
\let\@algcomment\relax
\newcommand\algcomment[1]{\def\@algcomment{\footnotesize#1}}
\renewcommand\fs@ruled{\def\@fs@cfont{\bfseries}\let\@fs@capt\floatc@ruled
  \def\@fs@pre{\hrule height.8pt depth0pt \kern2pt}%
  \def\@fs@post{}%
  \def\@fs@mid{\kern2pt\hrule\kern2pt}%
  \let\@fs@iftopcapt\iftrue}
\makeatother
\begin{document}

\title{NerfAcc: Efficient Sampling Accelerates NeRFs}

\author{Ruilong Li\\
UC Berkeley\\
{\tt\small ruilongli@berkeley.edu}
\and
Hang Gao\\
UC Berkeley\\
{\tt\small hangg@berkeley.edu}
\and
Matthew Tancik\\
UC Berkeley\\
{\tt\small tancik@berkeley.edu}
\and
Angjoo Kanazawa\\
UC Berkeley\\
{\tt\small kanazawa@berkeley.edu}
}

\twocolumn[{%
\renewcommand\twocolumn[1][]{#1}%
\maketitle

\begin{center}
\vspace{-0.5cm}
\includegraphics[width=\linewidth]{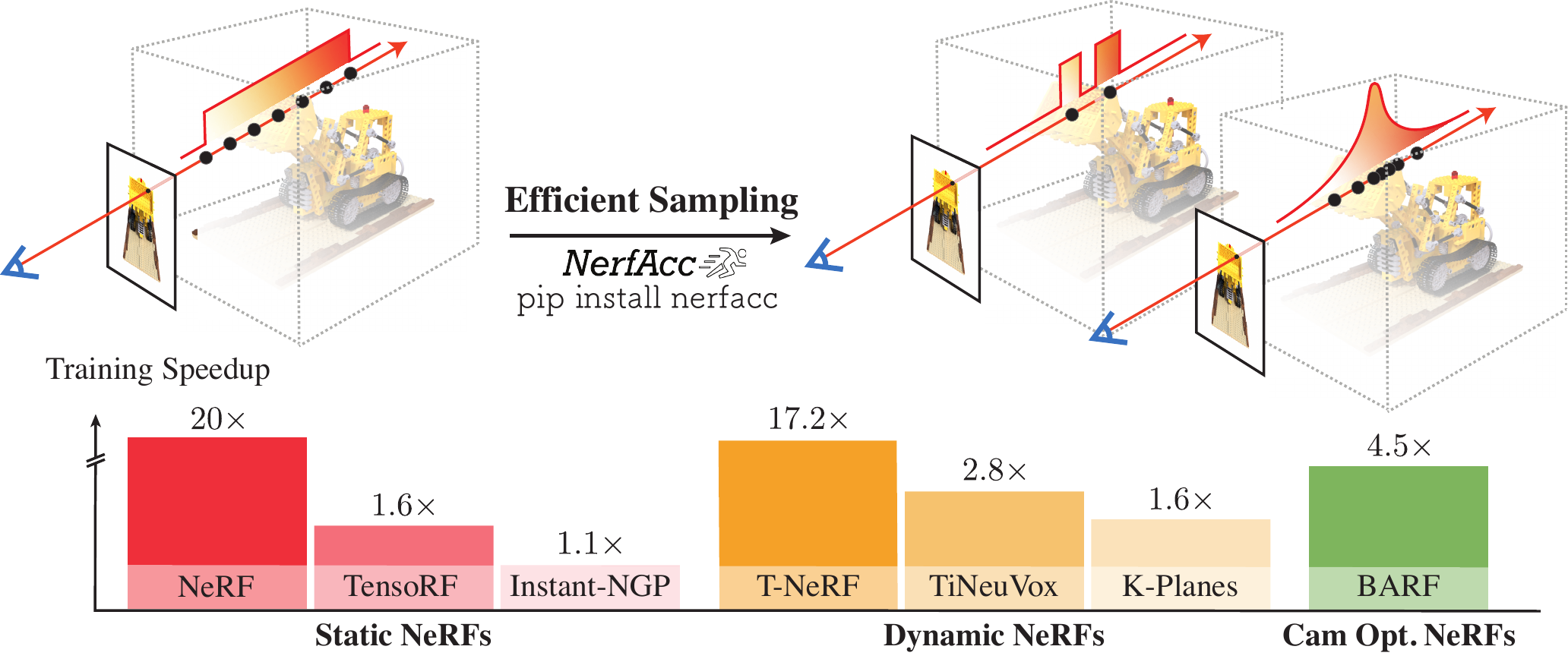}
\captionof{figure}{\textbf{NerfAcc Toolbox}. Our proposed toolbox, \emph{NerfAcc}, integrates advanced efficient sampling techniques that lead to significant speedups in training various recent NeRF papers with minimal modifications to existing codebases.}
\label{fig:teaser}
\end{center}

}]

\ificcvfinal\thispagestyle{empty}\fi


\begin{abstract}
   \looseness=-1 Optimizing and rendering Neural Radiance Fields is computationally expensive due to the vast number of samples required by volume rendering.
   Recent works have included alternative sampling approaches to help accelerate their methods, however, they are often not the focus of the work. 
   In this paper, we investigate and compare multiple sampling approaches and demonstrate that improved sampling is generally applicable across NeRF variants under an unified concept of transmittance estimator.
   To facilitate future experiments, we develop NerfAcc, a Python toolbox that provides flexible APIs for incorporating advanced sampling methods into NeRF related methods. 
   We demonstrate its flexibility by showing that it can reduce the training time of several recent NeRF methods by 1.5× to 20× with minimal modifications to the existing codebase. Additionally, highly customized NeRFs, such as Instant-NGP, can be implemented in native PyTorch using NerfAcc. Our code are open-sourced at \url{https://www.nerfacc.com}.

\end{abstract}

\section{Introduction}

Neural volume rendering has revolutionized the inverse rendering problem, with the Neural Radiance Field (NeRF)~\cite{mildenhall2021nerf} being a key innovation. The continuous radiance field representation allows for rendering novel views of a scene from any camera position. However, the optimization of a NeRF can be computationally expensive due to the neural representation of radiance field and the large number of samples required by volume rendering. These challenges have limited practical applications of NeRF-based optimization and rendering.

Several recent works have successfully reduced the computational cost of neural volume rendering by proposing more efficient radiance field representations~\cite{muller2022instant,yu2021plenoxels,yu2021plenoctrees,chen2022tensorf,sun2022direct,fridovich2023k}. While there are differences in the specific radiance field representations and their applications, most of these methods share a similar volume rendering pipeline which involves creating samples along the ray and accumulating them through alpha-composition. 

\looseness=-1
However, compared to the considerable efforts focused on developing efficient radiance field representations, there has been limited attention devoted to reducing the computational cost of neural volume rendering through \textit{efficient sampling}. While a few recent works have included alternative sampling approaches to accelerate their methods~\cite{muller2022instant,yu2021plenoxels,sun2022direct,barron2022mip360}, these methods are often not the main focus of the paper. Moreover, implementing advanced sampling approaches typically requires non-trivial efforts. For example, Instant-NGP~\cite{muller2022instant} and Plenoxels~\cite{yu2021plenoxels} both employ highly customized CUDA implementations to achieve spatial skipping during ray marching, which are tightly coupled with their respective radiance field implementations. Consequently, it can be challenging for researchers to benefit from these advanced sampling approaches in their own research.

In this paper, we investigate and compare several advanced sampling approaches from the literature and provide mathematical proofs demonstrating that they can all be viewed in a unified way of creating an \emph{estimation of transmittance} for importance sampling. Our analysis shows that by understanding the spectrum of sampling through transmittance estimator, novel sampling strategies can be explored. Based on this, we decouple the sampling procedure from the neural volumetric rendering pipeline and demonstrate that improved sampling is generally applicable across different variants of NeRF. Furthermore, to facilitate future experiments, we propose NerfAcc, a plug-and-play toolbox that provides a flexible Python API for integrating advanced sampling approaches into NeRF-related methods, ready for researchers to incorporate into their own codebases. We demonstrate that with less than $100$ lines of code change using NerfAcc, various NeRF methods~\cite{mildenhall2021nerf,chen2022tensorf,fang2022fast,fridovich2023k,lin2021barf,pumarola2021d,muller2022instant} can enjoy 1.5$\times$ to 20$\times$ training speedup with better performance. Notably, using the NerfAcc library, one can train an Instant-NGP~\cite{muller2022instant} model with pure Python code in the same amount of time, and achieve slightly better performance ($+0.2$dB) than the official pure CUDA implementation.

\looseness=-1 Our paper presents a unique contribution to the community. Unlike other papers that introduce novel algorithms, our work sheds light on the intricacies of various sampling approaches, which are often overlooked despite their significance. 
As far as we know, this work is the first paper that dives deep into this crucial aspect in the context of neural radiance field, and offer a novel, unified concept that allows researchers to view existing sampling algorithms in a fresh perspective. In addition, we provide a toolbox that facilitates research and development in this area. We believe translation of mathematical ideas into efficient, easy-to-modify implementation is fundamental to the research development. 
Overall, we hope that our proposed concept, along with the toolbox, can inspire new research ideas and accelerate progress in this field.

\section{Related Works}

\paragraph{NeRF Codebases.}
The recent explosion of NeRF-related research has led to numerous papers, many of which have released their own codebases~\cite{barron2021mip, barron2022mip360, chen2022tensorf, fridovich2023k, li2022tava, mildenhall2021nerf, park2021nerfies, yu2021pixelnerf, gao2022monocular}. These codebases address various tasks related to NeRF, including surface reconstruction~\cite{oechsle2021unisurf, wang2021neus, yariv2021volume}, radiance field representation~\cite{barron2021mip, barron2022mip360, muller2022instant, sun2022direct, yu2021plenoxels}, dynamic modeling~\cite{fang2022fast, fridovich2023k, li2022tava, pumarola2021d}, and camera optimization~\cite{lin2021barf, wang2021nerf}. However, each codebase is tailored to a specific task and supports only a single approach. While most of these methods share the same volume rendering pipeline of accumulating samples along the ray, transferring the implementation from one codebase to another requires non-trivial efforts. In this work, we address this common sampling problem by introducing NerfAcc, a plug-and-play toolbox that supports all the aforementioned tasks and methods, making it easy to integrate into any existing codebase.

\paragraph{NeRF Frameworks.}
Recently, several projects have been introduced to integrate different NeRF variants into a single framework, such as NeRF-Factory~\cite{nerf_factory}, Nerfstudio~\cite{tancik2023nerfstudio}, and Kaolin-Wisp~\cite{KaolinWispLibrary}. These frameworks have made significant progress in facilitating future NeRF-related research. NeRF-Factory offers a collection of NeRF variants~\cite{mildenhall2021nerf, zhang2020nerf++, sun2022direct, yu2021plenoxels, barron2021mip, barron2022mip360, verbin2022ref} with original implementations and focuses on comprehensive benchmarking. Nerfstudio consolidates critical techniques introduced in existing literature~\cite{mildenhall2021nerf, barron2021mip, barron2022mip360, muller2022instant, martin2021nerf} and provides modular components for the community to easily build on. Kaolin-Wisp builds upon the Kaolin~\cite{KaolinLibrary} framework and implements a set of voxel-based NeRF papers~\cite{takikawa2021neural, muller2022instant, takikawa2022variable}. However, these frameworks are designed to encourage researchers to develop within the framework, and do not benefit users working on their own codebases. Moreover, due to the high activity in the NeRF-related research, it is almost impossible to keep track of the latest developments and integrate advanced techniques into a single framework. Therefore, NerfAcc is designed as a standalone library that can be plugged into any codebase. It supports a wide range of NeRF related methods and can be easily maintained as new methods emerge.

\section{Importance Sampling via Transmittance}
\label{sec:method}

\begin{figure*}[!t]
\centering
\includegraphics[width=0.95\linewidth]{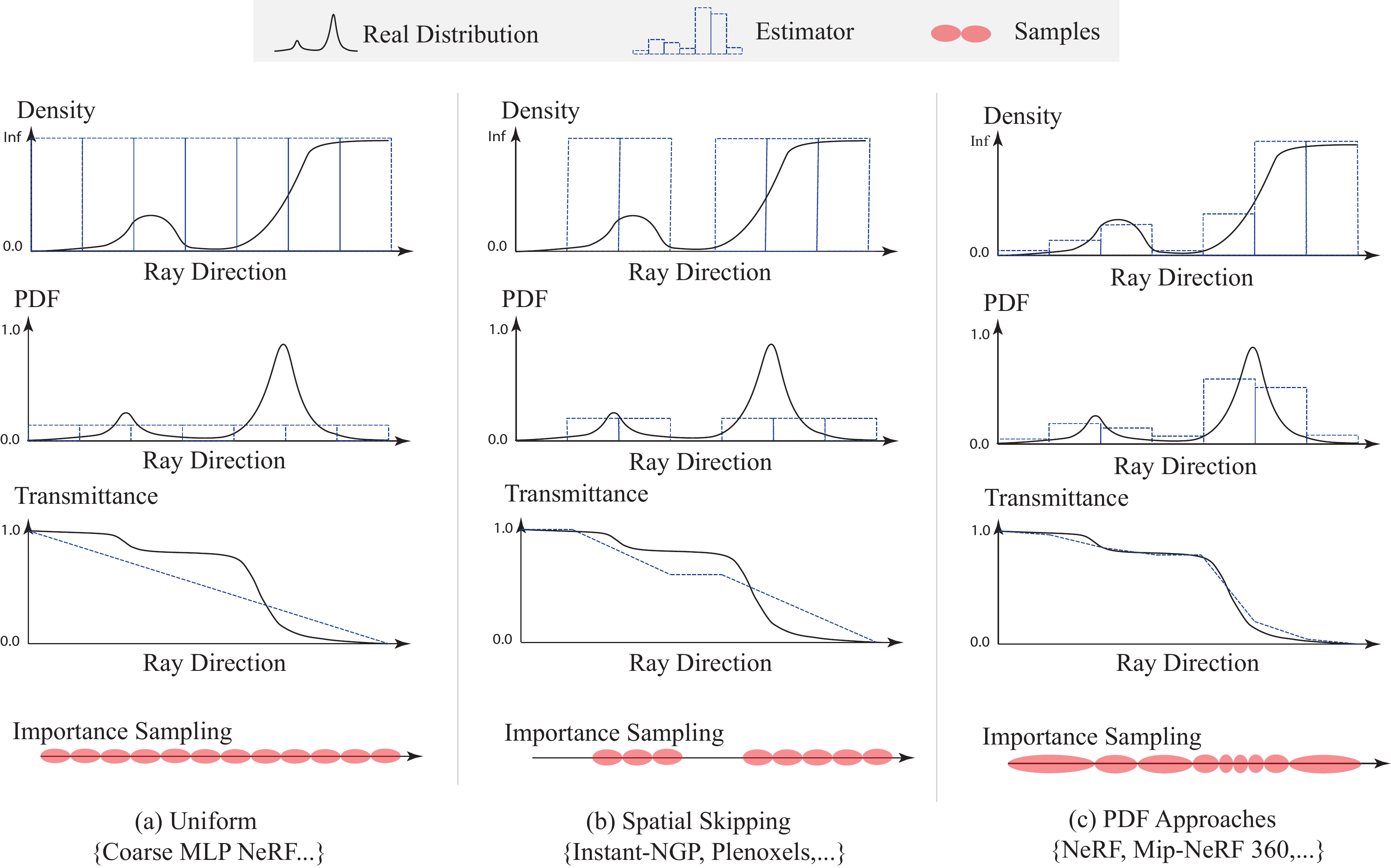}
\caption{\textbf{Illustration of Sampling via Transmittance Estimator.} Although spatial skipping approaches (e.g., Occupancy Grid in Instant-NGP~\cite{muller2022instant}) and PDF approaches (e.g., Proposal Network in Mip-NeRF 360~\cite{barron2022mip360}) appear distinct from each other, they can both be viewed as constructing a transmittance estimator from which samples can be created via importance sampling.
}
\label{fig:pdf}
\vspace{-0.5em}
\end{figure*}

\looseness=-1
Several advanced sampling approaches exist in the literature. For instance, Plenoxels~\cite{yu2021plenoxels} uses a sparse grid, Instant-NGP~\cite{muller2022instant} uses an occupancy grid, NeRF~\cite{mildenhall2021nerf} employs a coarse-to-fine strategy, Mip-NeRF 360~\cite{barron2022mip360} proposes proposal networks. However, creating a toolbox that can support all of these approaches is not an easy task since they operate in completely different ways. In this section, we demonstrate that these approaches can all be viewed in a unified way of importance sampling. We also provide a mathematical proof that \emph{transmittance is all you need for importance sampling}. Thus, each method essentially has its own way of creating an estimation of the transmittance along the ray, which we refer to as the \emph{transmittance estimator}. This observation enables us to unify different types of sampling approaches under the same framework and leads to our NerfAcc toolbox.

\subsection{Formulation}
\label{sec:formulation}

\looseness=-1
Efficient sampling is a well-explored problem in Graphics~\cite{fong2017production}, wherein the emphasis is on identifying regions that make the most significant contribution to the final rendering. This objective is generally accomplished through importance sampling, which aims to distribute samples based on the probability density function~(PDF), denoted as~$p(t)$, between the range of~$[t_n, t_f]$. 
By computing the cumulative distribution function~(CDF) through integration, \textit{i.e.}, $F(t) = \int_{t_n}^{t} p(v)\,dv$, samples are generated using the inverse transform sampling method:
\begin{equation}
t = F^{-1}(u) \quad \text{where} \quad u \sim \mathcal{U}[0,1].
\end{equation}

In volumetric rendering, the contribution of each sample to the final rendering is expressed by the accumulation weights $T(t)\sigma(t)$:
\begin{equation}
\begin{gathered}
C(\mathbf{r}) = \int_{t_n}^{t_f} T(t)\,\sigma(t)\,c(t)\,dt
\\
\quad \text{where} \quad
T(t) = \exp\left(-\int_{t_n}^{t}\,\sigma(s)\,ds\right).
\end{gathered}
\end{equation}
Hence, the PDF for volumetric rendering is $p(t) = T(t)\sigma(t)$ and the CDF $F(t) = \int_{t_n}^{t} p(v)\,dv$ can be derived as a function of transmittance $T(t)$:
\begin{equation}
\begin{aligned}
F(t) & = \int_{t_n}^{t} T(v)\sigma(v)\,dv \\
& = \int_{t_n}^{t} \frac{d}{dv} \left[ - \exp \left(-\int_{t_n}^{v}\sigma(s)ds \right) \right]dv \\
& = 1 - \exp \left(-\int_{t_n}^{t}\sigma(s)ds \right) \\
& = 1 - T(t).
\end{aligned}
\end{equation}

\begin{table*}[t]
    \centering
    \tablestyle{2pt}{1.05}
    \begin{tabular}{y{65}x{110}x{110}x{60}x{80}}
        \toprule
        \textbf{Sampling Methods} & \textbf{PDF Estimator $p(t_i)$} & \textbf{Transmittance Estimator $T(t_i)$} & \textbf{Update $\mathcal{F}$} & \textbf{Instantiation} \\
        
        \midrule
        Uniform & 
        Constant & 
        $1-\frac{t-t_n}{t_f-t_n}$ & 
        - & 
        Coarse MLP NeRF~\cite{mildenhall2021nerf}  \\
        
        \arrayrulecolor{black!20}\midrule
        \multirow{2}{*}{Spatial Skipping} & \multirow{2}{*}
        {$\frac{\mathbb{1}\big[\sigma(t_i) > \tau\big]}{\sum_{j=1}^{n} \mathbb{1}\big[\sigma(t_i) > \tau\big]}$} & 
        \multirow{2}{*}{$1 - \frac{\sum_{j=1}^{i-1}\mathbb{1}\big[\sigma(t_i) > \tau\big]}{\sum_{j=1}^{n} \mathbb{1}\big[\sigma(t_i) > \tau\big]}$} & 
        EMA & Instant-NGP~\cite{muller2022instant}  \\
        {} & {} & {} & SGD & Plenoxels~\cite{yu2021plenoxels}  \\
        
        \arrayrulecolor{black!20}\midrule
        \multirow{2}{*}{PDF Approaches} & \multirow{2}{*}{$\sigma(t_i)\exp(-\sigma(t_i)\,dt)$} & \multirow{2}{*}{$\exp(-\sum_{j=1}^{i-1}\sigma(t_i)\,dt)$} & SGD & Fine MLP NeRF~\cite{mildenhall2021nerf}  \\
        {} & {} & {} & SGD & Mip-NeRF 360~\cite{barron2022mip360}  \\

        \arrayrulecolor{black}\bottomrule
    \end{tabular}
    \caption{
    \textbf{Mathematical Formulations of Different Sampling Approaches.} We outlines the significant mathematical distinctions among each sampling approach, under the perspective of transmittance estimator. See Section~\ref{sec:formulation} for notations.
    }
    \label{tab:volrend_compare}
\vspace{-0.5em}
\end{table*}

Therefore, inverse sampling the CDF $F(t)$ is equivalent to inverse sampling the transmittance $T(t)$. 
Thus a transmittance estimator is sufficient to determine the optimal samples. 
Intuitively, this suggests to put more samples around regions where the transmittance changes rapidly -- and that is exactly what happens when a ray hit a surface. Implementation wise, this observation enables us to compute the CDF directly using $1 - T(t)$, instead of the computationally expensive integral $\int_{t_n}^{t} T(v)\sigma(v)\,dv$, which is the standard implementation adopted by many popular codebases~\cite{mildenhall2021nerf,barron2021mip,barron2022mip360,tancik2023nerfstudio,fridovich2023k}.

\looseness=-1
While advanced transmittance estimator techniques, such as delta tracking~\cite{von195113} with Monte Carlo sampling, are utilized in production-level volumetric rendering in Graphics~\cite{fong2017production}, NeRFs operate in a distinct setting in which the scene geometry is not predefined but optimized on the fly.
During NeRF optimization, the radiance field changes between iterations, necessitating the dynamic update of the transmittance estimator at each step~$k$:
\begin{equation}
    \mathcal{F}: T(t)^{k-1} \mapsto T(t)^{k}.
\end{equation}
This introduces additional challenges to efficient sampling because it becomes more difficult to accurately estimate the transmittance from a radiance field that is constantly changing. Current approaches employ either exponential moving average (EMA) or stochastic gradient descent (SGD) as the update function $\mathcal{F}$. However, we note that there may be other update functions that could be explored.

With these concepts in mind, let us now examine some of the existing approaches towards efficient sampling. 

\paragraph{Uniform.} 
If a transmittance estimator is not available, the only assumption we can make is that every point along the ray contributes equally to the final rendering. 
Mathematically, this assumption translates to a constant PDF and a linearly decaying transmittance $T(t) = 1-(t-t_n)/(t_f-t_n)$. 
In this case, the sampling process is equivalent to uniformly sampling along the ray, i.e., $t_i = t_n + (t_f - t_n) \cdot u_i$. 
It is worth noting that every NeRF model that uses uniform sampling inherently assumes this linear transmittance decay, such as the coarse level in vanilla NeRF~\cite{mildenhall2021nerf}. 
See Fig.~\ref{fig:pdf}(a) for the illustration.

\paragraph{Spatial Skipping.} 
A more sophisticated approach to improve uniform sampling is to identify empty regions and skip them during sampling, as proposed in Instant-NGP's Occupancy Grid~\cite{muller2022instant} and Plenoctrees' Sparse Grid~\cite{yu2021plenoctrees}. This technique binarizes the density along the ray with a conservative threshold $\tau$: $\hat{\sigma}(t_i) = \mathbb{1}\big[\sigma(t_i) > \tau\big]$. Consequently, the piece-wise constant PDF can be expressed as $p(t_i) = \hat{\sigma}(t_i) / \sum_{j=1}^{n} \hat{\sigma}(t_j)$, and the piece-wise linear transmittance estimator is $T(t_i) = 1 - \sum_{j=1}^{i-1}\hat{\sigma}(t_j) / \sum_{j=1}^{n} \hat{\sigma}(t_j)$. 
To update this estimator during optimization, Instant-NGP~\cite{muller2022instant} directly updates the cached density with exponential moving average (EMA) over iteration~$k$: $\sigma(t_i)^{k} = \gamma \cdot \sigma(t_i)^{k-1} + (1 - \gamma) \cdot \sigma(t_i)^{k}$. 
Meanwhile, Plenoxels~\cite{yu2021plenoxels} updates the density via the gradient descent through the rendering loss. 
See Fig.~\ref{fig:pdf}(b) for an illustration.

\paragraph{PDF Approaches.} 
Another type of approach is to directly estimate the PDF along the ray with discrete samples. In vanilla NeRF~\cite{mildenhall2021nerf}, the coarse MLP is trained using volumetric rendering loss to output a set of densities ${\sigma(t_i)}$. This allows for the creation of a piece-wise constant PDF: $p(t_i) = \sigma(t_i)\exp(-\sigma(t_i)\,dt)$, and a piece-wise linear transmittance estimator $T(t_i) = \exp(-\sum_{j=1}^{i-1}\sigma(t_i)\,dt)$. This approach was further improved in Mip-NeRF 360~\cite{mildenhall2021nerf} with a PDF matching loss, which allows for the use of a much smaller MLP in the coarse level, namely Proposal Network, to speedup the PDF construction. 
In both cases, the transmittance estimator is updated through gradient descent.
See Fig.~\ref{fig:pdf}(c) for an illustration.

Table~\ref{tab:volrend_compare} provides a mathematical summary and comparison of these approaches. Additionally, we present an illustration in Figure~\ref{fig:pdf} to provide an intuitive comparison of these approaches from the perspective of PDF (second row) and transmittance (third row), as well as how the samples can be created from the transmittance estimator via importance sampling (last row). This visualization also reveals some pros and cons for each approach, which we will discuss in in Section.~\ref{sec:discussions}.

\subsection{Design Spaces}
\label{sec:design}
\paragraph{Choice of Representations.} The transmittance estimator can use either an explicit voxel~\cite{muller2022instant,yu2021plenoxels,chen2022tensorf}, an MLP~\cite{barron2021mip,barron2022mip360,mildenhall2021nerf}, or a hybrid representation~\cite{tancik2023nerfstudio}. Depending on whether the estimator is explicit or not, it can be updated with either rule-based EMA~\cite{muller2022instant,chen2022tensorf} or gradient descent with some supervision~\cite{yu2021plenoxels,barron2022mip360,mildenhall2021nerf,tancik2023nerfstudio}. Generally, voxel-based estimators are faster than implicit (e.g., MLP-based) estimators but suffer more from aliasing issues. It is worth noting that the transmittance estimator's representation can significantly benefit from advances in radiance field representation. For example, the Nerfacto model~\cite{tancik2023nerfstudio} uses the most recent hybird-representation HashEncoding~\cite{muller2022instant} for both the radiance field and the sampling module, achieving the best quality-speed tradeoff in in-the-wild settings.

\paragraph{Handling Unbounded Scenes.} So far, we have only discussed sampling within a bounded area $[t_n, t_f]$. For unbounded scenes, it is impossible to densely sample along the ray. Similar to the mipmaps used in graphics rendering, a general solution is to sample more coarsely as the ray goes further, as objects farther away appear in fewer pixels on the image plane. This can be achieved by creating a bijective mapping function $\Phi: s \in [s_n, s_f] \mapsto t\in [t_n, +\infty]$, and performing sampling in the $s$-space instead of the $t$-space. Several papers~\cite{barron2022mip360,zhang2020nerf++,reiser2023merf} that work on unbounded scenes have introduced different mapping functions $\Phi$, to which we refer our readers for details.

\subsection{Discussions}
\label{sec:discussions}
\paragraph{Pros and Cons.} Sampling with uniform assumption is the easiest one to implement but with lowest efficiency in most cases. 
Spatial skipping is a more efficient technique since most of the 3D space is empty, but it still samples uniformly within occupied but occluded areas that contribute little to the final rendering (e.g., the last sample in Figure~\ref{fig:pdf}(b)).
PDF-based estimators generally provide more accurate transmittance estimation, enabling samples to concentrate more on high-contribution areas (e.g., surfaces) and to be more spread out in both empty and occluded regions. However, this also means that samples are always spread out throughout the entire space without any skipping, as shown in Figure~\ref{fig:pdf}(c).
Moreover, the current approaches all introduce aliasing effects to the volumetric rendering due to either (1) the piece-wise linearity assumption for estimating transmittance along the ray~\cite{barron2021mip, barron2022mip360, mildenhall2021nerf,yu2021plenoxels,muller2022instant} as illustrated in Figure~\ref{fig:pdf}, or (2) the underlying voxel representation for the transmittance estimator~\cite{yu2021plenoxels,muller2022instant} as discussed in Section~\ref{sec:design}. A recent work,  Zip-NeRF~\cite{barron2023zipnerf}, addresses the aliasing problem tied to these two exact issues (called ``z-aliasing'' and ``xy-aliasing'' in their work), which are naturally revealed under our unified framework.

\paragraph{Implementation Difficulties.} The current implementations for efficient sampling are all highly customized and tightly integrated with the specific radiance field proposed in each paper. For instance, spatial skipping is implemented with customized CUDA kernels in Instant-NGP \cite{muller2022instant} and Plenoxels \cite{yu2021plenoxels}. Mip-NeRF 360 \cite{barron2022mip360}, K-planes \cite{fridovich2023k}, and Nerfacto \cite{tancik2023nerfstudio} have a proposal network implemented but it is closely integrated with their repositories and can only support limited types of radiance fields that come with the repository. However, as shown before, the sampling process is independent of the radiance field representation, thus it should be easily transferable across different NeRF variants. Due to the various implementation details, it typically requires significant effort to correctly implement an efficient sampling approach from scratch. Therefore, having an implementation that is easily transferable from repository to repository would be valuable in supporting future research on NeRF.

\paragraph{Insights from Unified Formulation.} Comprehending the sampling spectrum through the lens of Transmittance Estimator paves the way for investigating novel sampling strategies. For example, our framework reveals that the Occupancy Grid from Instant-NGP~\cite{muller2022instant} and the Proposal Network from Mip-NeRF 360~\cite{barron2022mip360} are not mutually exclusive but complementary, as both aim to estimate the transmittance along the ray. Therefore, combining them becomes straight-forward: one can first compute the transmittance with the occupancy grid, and then refine the estimated transmittance with a proposal network. This enables both skipping on the empty space and concentrating the samples onto the surface. We explore this approach in Section~\ref{ref:combined_smapling} and demonstrate that it overcomes the limitation of the proposal network approach, which always samples the entire space. Furthermore, this formulation could potentially shed light on questions such as how to enhance the sampling procedure with depth information or other priors, which we encourage readers to investigate further.

\section{NerfAcc Toolbox}

\newcommand\pythoninline[1]{{\pythonstyle\lstinline!#1!}}

\begin{algorithm}[t]
\caption{\small \hspace{-0.5em} \textbf{: NerfAcc Rendering Pipeline.} In NerfAcc, the samples are created from a transmittance estimator, which can be updated during the NeRF training. See Section~\ref{sec:impl-details} for details.}
\label{alg:code}
\definecolor{codeblue}{rgb}{0.25,0.5,0.5}
\definecolor{codekw}{rgb}{0.85, 0.18, 0.50}
\begin{lstlisting}[language=python]
# nerf: a radiance field model.
# r_o: ray origins. (n_rays, 3)
# r_d: ray normalized directions. (n_rays, 3)

# e.g., Prop Net, Occ Grid.
estimator = nerfacc.TransmittanceEstimator()

def density_fn(t0, t1, r_id):
    """Query density from nerf."""
    return nerf.density(r_o[r_id], r_d[r_id], t0, t1)

def rgb_density_fn(t0, t1, r_id):
    """Query rgb and density from nerf."""
    return nerf.forward(r_o[r_id], r_d[r_id], t0, t1)

# Efficient sampling.
# (t0, t1, r_id): packed samples. (all_samples,)
t0, t1, r_id = nerfacc.sampling(
    r_o, r_d, estimator, density_fn=density_fn
)
# Differentiable volumetric rendering. 
color, opacity, depth, aux = nerfacc.rendering(
   t0, t1, r_id, rgb_density_fn=rgb_density_fn
)
# Update the transmittance estimator.
estimator.update_every_n_steps(t0, t1, r_id, aux)
# nerf, r_o and r_d all receive gradients.
F.mse_loss(color, color_gt).backward()
\end{lstlisting}
\end{algorithm}
In this paper, we present \emph{NerfAcc} toolbox, designed for \textbf{Ne}ural \textbf{r}adiance \textbf{f}ield \textbf{Acc}eleration. It provides efficient sampling for volumetric rendering, that is universally applicable and easily integrable for a diverse range of radiance fields~\cite{mildenhall2021nerf,chen2022tensorf,muller2022instant,lin2021barf,wang2021neus}. In this section, we first introduce the design principles of this toolbox, along with critical implementation details. To demonstrate its flexibility, we further show that it can significantly speedup the training of various NeRF-related papers by $1.5\times$ to $20\times$ with only minor modifications to existing codebases.

\subsection{Design Principles}

This library is designed with these goals in mind:
\begin{itemize}[topsep=4pt,itemsep=2pt,partopsep=2pt,parsep=2pt]
\item\textbf{Plug-and-play.} Our primary objective is to ease the challenges of implementing an efficient volumetric sampling technique for the research community. Therefore, NerfAcc is designed as a standalone library that can be easily installed from PyPI on both Windows and Linux platforms, and seamlessly integrated into any PyTorch codebase.

\item\textbf{Efficiency \& Flexibility.} To maximize the speed of the code, we fuse the operations into CUDA kernels as much as possible, while exposing flexible Python APIs to the users.

\item\textbf{Radiance Field Complexity.}  We target on supporting any radiance fields that are designed for per-scene optimization, including both density-based and SDF-based fields, for both static and dynamic scenes.

\end{itemize}

\providecommand\animage{}
\renewcommand{\animage}[2]{
    \frame{\includegraphics[width=\linewidth,clip,trim=#1]{figures/assets/qualitatives/#2}}
}
\providecommand\textimage{}
\renewcommand{\textimage}[4]{
	\frame{\begin{overpic}[width=\linewidth,clip,trim=#1]{figures/assets/qualitatives/#2}\put(0,945){\footnotesize\sethlcolor{black}\textcolor{white}{\hl{#3$/$#4}}}\end{overpic}}
}
\begin{figure*}[t!]
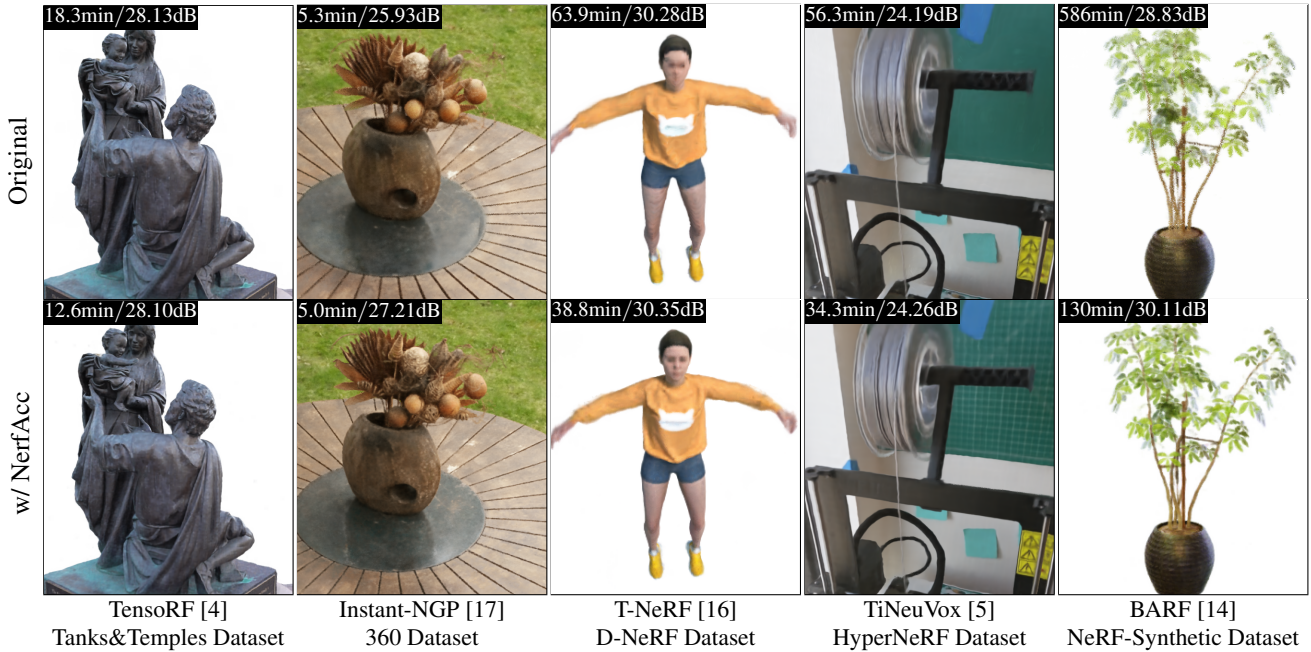

    \centering
    \begin{minipage}{\linewidth}
        \setlength{\tabcolsep}{1pt}
        \renewcommand{\arraystretch}{0.1}
        \begin{tabularx}{\textwidth}{@{}c*{5}{C}@{}}
            \makebox[20pt]{\raisebox{50pt}{\rotatebox[origin=c]{90}{\text{Original}}}}\hspace{-3pt} &
            \textimage{541 116 630 80}{tnt_tensorf/Family_12_orig.png}{18.3min}{28.13dB} &
            \textimage{457 370 510 80}{360_ngp/garden_18_orig.png}{5.3min}{25.93dB} &
            \textimage{70 60 60 20}{dnerf_kplanes/jumpingjacks_4_orig.png}{63.9min}{30.28dB} &
            \textimage{60 396 0 0}{hypernerf_tineuvox/3dprinter_86_orig.png}{56.3min}{24.19dB} &
            \textimage{68 60 62 20}{noisy_blender_barf/ficus_26_orig.png}{586min}{28.83dB}
            \\
            \makebox[20pt]{\raisebox{50pt}{\rotatebox[origin=c]{90}{\text{w/ NerfAcc}}}}\hspace{-3pt} &
            \textimage{541 116 630 80}{tnt_tensorf/Family_12_occ.png}{12.6min}{28.10dB} &
            \textimage{457 370 510 80}{360_ngp/garden_18_prop.png}{5.0min}{27.21dB} &
            \textimage{70 60 60 20}{dnerf_kplanes/jumpingjacks_4_occ.png}{38.8min}{30.35dB} &
            \textimage{60 396 0 0}{hypernerf_tineuvox/3dprinter_86_prop.png}{34.3min}{24.26dB} &
            \textimage{68 60 62 20}{noisy_blender_barf/ficus_26_occ.png}{130min}{30.11dB}
            \\
            [2pt] &
            {\small TensoRF~\cite{chen2022tensorf}} &
            {\small Instant-NGP~\cite{muller2022instant}} &
            {\small T-NeRF~\cite{mildenhall2021nerf}} &
            {\small TiNeuVox~\cite{fang2022fast}} &
            {\small BARF~\cite{lin2021barf}} 
            \\
             & & & & &
            \\
            [2pt] &
            {\small Tanks\&Temples Dataset} &
            {\small 360 Dataset} &
            {\small D-NeRF Dataset} &
            {\small HyperNeRF Dataset} &
            {\small NeRF-Synthetic Dataset} 
            \\
        \end{tabularx}
    \end{minipage}
    \vspace{-4pt}
    \caption{
        \textbf{Qualitative Results.} NerfAcc is able to significantly reduce the training time of various NeRF-related methods across multiple datasets, while also yielding slightly improved quality in the majority of cases. The training time and test PSNR are shown on the left corner of each image.
    }
    \label{fig:qualitatives}
\vspace{-0.5em}
\end{figure*}

\begin{table}
\begin{subtable}{\linewidth}
\centering
    \tablestyle{2pt}{1.05}
    \resizebox{\linewidth}{!}{%
    \begin{tabular}{y{63}x{28}x{28}x{34}x{28}x{28}}
        \toprule
        \textbf{Methods} & \textbf{Dataset} & \textbf{Speedup} & \textbf{$\mathbf{T}$ (min)}~$\downarrow$ & \textbf{PSNR}~$\uparrow$ & \textbf{LPIPS}~$\downarrow$ \\

        \midrule
        TensoRF~\cite{chen2022tensorf} & \multirow{2}{*}{T\&T} & \multirow{2}{*}{$\mathbf{1.5\times}$} & 18.3 & \hlg{28.13} & \hlg{0.143}\\
        \hspace{0.5em} + nerfacc (occ) & & & \hlg{12.6} & 28.10 & 0.150\\
        
        \arrayrulecolor{black!20}\midrule
        TensoRF~\cite{chen2022tensorf} & \multirow{2}{*}{Syn.} & \multirow{2}{*}{$\mathbf{1.6\times}$} & 10.6 & \hlg{32.52} & 0.047\\
        \hspace{0.5em} + nerfacc (occ) & & & \hlg{6.5} & 32.51 & \hlg{0.044}\\
        
        \arrayrulecolor{black!20} \arrayrulecolor{black!20}\midrule
        NeRF~\cite{mildenhall2021nerf} & \multirow{2}{*}{Syn.} & \multirow{2}{*}{$\mathbf{20\times}$} & $>$1000 & 31.00 & 0.081\\
        \hspace{0.5em} + nerfacc (occ)$^\dag$ & & & \hlg{52.0} & \hlg{31.55} & \hlg{0.072}\\
        
        \arrayrulecolor{black!20}\midrule
        Instant-NGP~\cite{muller2022instant} & \multirow{3}{*}{Syn.} & \multirow{3}{*}{$\mathbf{1.0\times}$} & \hlg{4.4} & 32.35 & -\\
        \hspace{0.5em} + nerfacc (occ)$^\dag$ & & & \hlg{4.4} & \hlg{32.55} & \hlg{0.056}\\
        \hspace{0.5em} + nerfacc (prop)$^\dag$ & & & 5.2 & 31.40 & 0.064\\
        
        \arrayrulecolor{black!20}\midrule
        Instant-NGP~\cite{muller2022instant} & \multirow{3}{*}{360} & \multirow{3}{*}{$\mathbf{1.1\times}$} & 5.3 & 25.93 & -\\
        \hspace{0.5em} + nerfacc (occ)$^\dag$ & & & 5.0 & 26.41 & 0.353\\
        \hspace{0.5em} + nerfacc (prop)$^\dag$ & & & \hlg{4.9} & \hlg{27.58} & \hlg{0.292}\\
        \arrayrulecolor{black!}\bottomrule
    \end{tabular}
    }
    \vspace{-0.5em}
    \caption{\textbf{Static NeRFs.}}
    \label{tab:result_static}
\end{subtable}

\vspace{0.5em}
\begin{subtable}{\linewidth}
    \centering
    \tablestyle{2pt}{1.05}
    \resizebox{\linewidth}{!}{%
    \begin{tabular}{y{63}x{28}x{28}x{34}x{28}x{28}}
        \toprule
        \textbf{Methods} & \textbf{Dataset} & \textbf{Speedup} & \textbf{$\mathbf{T}$ (min)}~$\downarrow$ & \textbf{PSNR}~$\uparrow$ & \textbf{LPIPS}~$\downarrow$  \\

        \midrule
        TiNeuVox~\cite{park2021hypernerf} & \multirow{3}{*}{Hyper.} & \multirow{3}{*}{$\mathbf{1.7\times}$} & 56.3 & 24.19 & 0.425\\
        \hspace{0.5em} + nerfacc (occ) & & & \hlg{33.0} & 24.19 & 0.434\\
        \hspace{0.5em} + nerfacc (prop) & & & 34.3 & \hlg{24.26} & \hlg{0.398}\\

        \arrayrulecolor{black!20}\midrule
        TiNeuVox~\cite{fang2022fast} & \multirow{2}{*}{D-NeRF} & \multirow{2}{*}{$\mathbf{2.8\times}$} & 11.8 & 31.14 & 0.050\\
        \hspace{0.5em} + nerfacc (occ) & & & \hlg{4.2} & \hlg{31.75} & \hlg{0.038}\\

        \arrayrulecolor{black!20}\midrule
        K-Planes~\cite{fridovich2023k} & \multirow{2}{*}{D-NeRF} & \multirow{2}{*}{$\mathbf{1.6\times}$} & 63.9 & 30.28 & 0.043\\
        \hspace{0.5em} + nerfacc (occ) & & & \hlg{38.8} & \hlg{30.35} & \hlg{0.042}\\
        
        \arrayrulecolor{black!20}\midrule
        T-NeRF~\cite{pumarola2021d} & \multirow{2}{*}{D-NeRF} & \multirow{2}{*}{$\mathbf{20\times}$} & $>$1000 & 28.78 & 0.069\\
        \hspace{0.5em} + nerfacc (occ)$^\dag$ & & & \hlg{58.0} & \hlg{32.22} & \hlg{0.040}\\
        
        \arrayrulecolor{black!}\bottomrule
    \end{tabular}
    }
    \vspace{-0.5em}
    \caption{\textbf{Dynamic NeRFs.}}
    \label{tab:result_dynamic}
\end{subtable}

\vspace{0.5em}
\begin{subtable}{\linewidth}
    \centering
    \tablestyle{2pt}{1.05}
    \resizebox{\linewidth}{!}{%
    \begin{tabular}{y{53}x{24}x{30}x{34}x{25}x{25}x{38}}
        \toprule
        \textbf{Methods} & \textbf{Dataset} & \textbf{Speedup} & \textbf{$\mathbf{T}$ (min)}~$\downarrow$ & \textbf{PSNR}~$\uparrow$ & \textbf{LPIPS}~$\downarrow$ & $\mathbf{E_R}$ / $\mathbf{E_T}$~$\downarrow$ \\

        \midrule
        BARF~\cite{park2021hypernerf} & \multirow{2}{*}{Syn.} & \multirow{2}{*}{$\mathbf{4.5\times}$} & 586 & 28.83 & 0.054 & 0.19 / 0.74\\
        \hspace{0.5em} + nerfacc (occ) & & & \hlg{130} & \hlg{30.11} & \hlg{0.044} & \hlg{0.07 / 0.35}\\

        \bottomrule
    \end{tabular}
    }
    \vspace{-0.5em}
    \caption{\textbf{NeRFs for Camera Optimization.} Camera rotational / transnational errors are denoted as $E_R/E_T$. 
    $E_T$ is scaled by 100.}
    \label{tab:result_camera}

\end{subtable}

\caption{\textbf{Improvements on Various NeRFs.} Experiments are conducted by replacing necessary code in the official repositories, except for those marked by $^\dag$ that are based on our re-implementations.}
\vspace{-0.5em}
\label{tab:results}
\end{table}
\begin{figure}[!t]
\centering
\includegraphics[width=\linewidth]{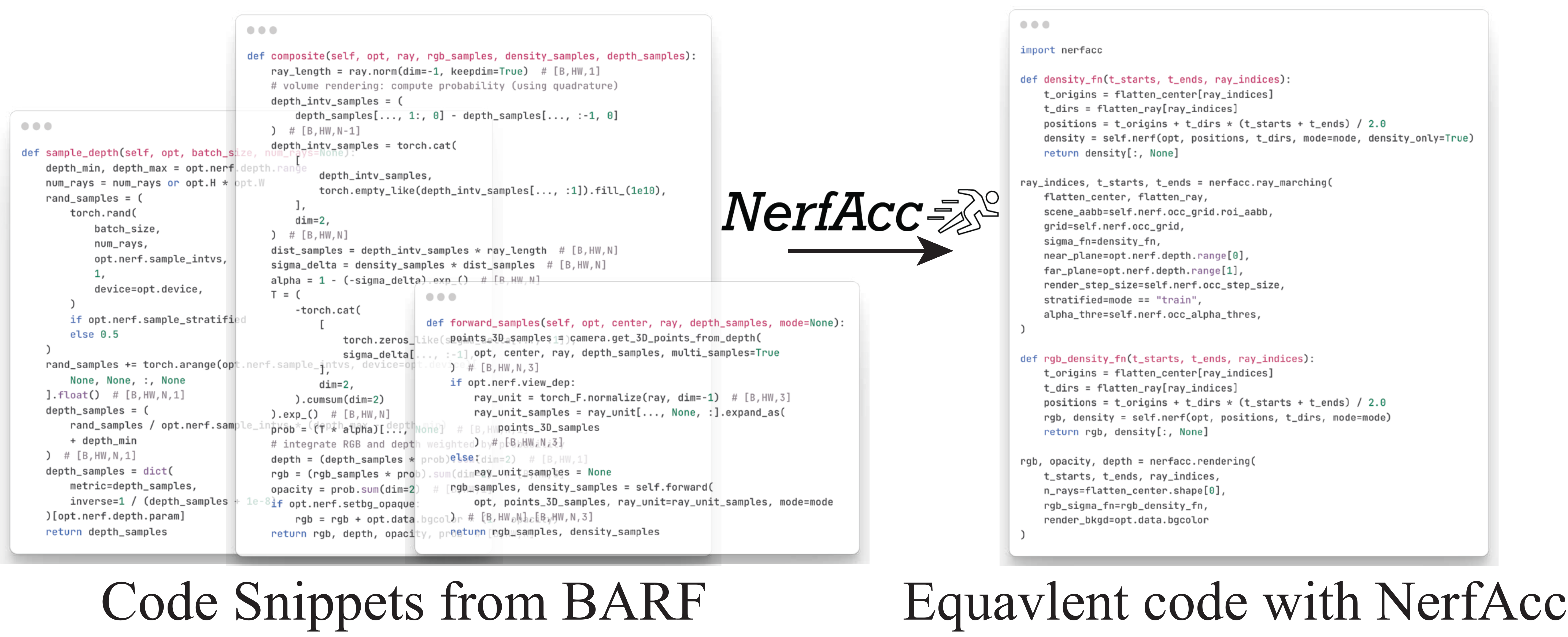}
\caption{\textbf{Plug-and-play Example in BARF~\cite{lin2021barf}'s Repository.} With around $50$ lines of code change, both training speed and performance can be greatly improved with efficient sampling equipped in NerfAcc, as shown in Table~\ref{tab:result_dynamic}.
}
\vspace{-0.5em}
\label{fig:git_diff}
\end{figure}

\subsection{Implementation Details}
\label{sec:impl-details}

NerfAcc incorporates two advanced sampling methods that can be decoupled from the radiance field representation, namely the Occupancy Grid from Instant-NGP~\cite{muller2022instant} and the Proposal Network from Mip-NeRF 360~\cite{barron2022mip360}. Algorithm~\ref{alg:code} presents the pseudo code for volumetric rendering using the NerfAcc toolbox. In this section, we do not dive into the details of each algorithm, as we basically follow the original paper's implementation. Instead, we discuss the implementation designs that are crucial to maintain high efficiency and flexibility of this toolbox.

\paragraph{Sample as Interval.} 
In the NerfAcc toolbox, 
instead of representing each sample with a coordinate $x$, we represent the sample as an interval along the ray $(t_0, t_1, r)$, where $t_0$ and $t_1$ are the start and end of the interval along the $r$-th ray. This interval-based representation offers three key advantages.
Firstly, representing a sample as an interval instead of a single point allows us to support research based on cone-based rays for anti-aliasing, such as Mip-NeRF~\cite{barron2021mip} and Mip-NeRF 360~\cite{barron2022mip360}. 
Secondly, since in almost all cases $t_i$ does not require gradients, using $(t_0, t_1, r)$ instead of $(\mathbf{x}_0, \mathbf{x}_1)$ to represent the interval allows for the detachment of the sampling process from the differentiable computational graph, thereby maximizing its speed. 
Lastly, the ray id $r$ attached to each sample enables support for various numbers of samples across rays with a packed tensor, which we will discuss in the next paragraph\textbf{}. 
A similar representation has been adopted in Nerfstudio~\cite{tancik2023nerfstudio} to support various radiance fields.

\paragraph{Packed Tensor.} To support sampling with spatial skipping, it is necessary to consider that each ray may result in a different number of valid samples. Storing the data as a tensor with shape $\texttt{(n\_rays, n\_samples, ...)}$ and an extra mask with shape \texttt{(n\_rays, n\_samples)} to indicate which samples are valid leads to significant inefficient memory usage when large portions of space are empty. To address this, in NerfAcc, we represent samples as ``packed tensors'' with shape \texttt{(all\_samples, ...)}, in which only valid samples are stored (see Algo.~\ref{alg:code}). To keep track of the associated rays for each sample, we also host an integer tensor with shape \texttt{(n\_rays, 2)}, which stores the start index in the packed tensors and the number of samples on this ray. This approach is similar to that used in Instant-NGP~\cite{muller2022instant} and PyTorch3D~\cite{ravi2020accelerating}.

\paragraph{No Gradient Filtering.} After importance sampling, inaccurate transmittance estimations can result in some samples lying in empty or occluded spaces, particularly in spatial skipping methods like Occupancy Grid. These samples can be filtered before being included in PyTorch's differentiable computation graph by evaluating their transmittance using the radiance field with gradients disabled. As backward passes are not required during filtering, this is much faster (${\sim}10\times$) than keeping all samples in the computation graph. In practice, samples with transmittance below $10^{-4}$ are disregarded during this process with almost no influence on the rendering quality. Note that this strategy is inspired from Instant-NGP~\cite{muller2022instant}'s implementation.

\subsection{Case Studies}

We showcase the flexibility of NerfAcc on three types of NeRFs across seven papers: static NeRFs (NeRF~\cite{mildenhall2021nerf}, TensoRF~\cite{chen2022tensorf}, Instant-NGP~\cite{muller2022instant}); dynamic NeRFs (D-NeRF~\cite{pumarola2021d}, K-Planes~\cite{fridovich2023k} TiNeuVox~\cite{fang2022fast}); and a NeRF variation for camera optimization (BARF~\cite{lin2021barf}). Although many of these methods, \textit{e.g.}, Instant-NGP, TensoRF, TiNeuVox and K-Planes, have already been highly optimized for efficiency, we are still able to accelerate their training by a large margin and achieve slightly better performance on nearly all cases. It is worth mentioning that experiments with TensoRF, TiNeuVox, K-Planes and BARF are conducted by integrating NerfAcc into the \textit{official codebase} with around 100 lines of code change. The results of our experiments, including those of our baselines, are presented in Table~\ref{tab:result_static},\ref{tab:result_dynamic}, and\ref{tab:result_camera}, all of which were conducted under the same physical environment using a single NVIDIA RTX A5000 GPU to facilitate comparison, as per~\cite{sun2022improved}.
Aside from the experiments reported in this paper, NerfAcc has also been integrated into a few popular open-source projects such as \texttt{nerfstudio}~\cite{tancik2023nerfstudio} for density-based NeRFs, as well as \texttt{sdfstudio}~\cite{Yu2022SDFStudio} and \texttt{instant-nsr-pl}~\cite{instant-nsr-pl} for SDF-based NeRFs.

\paragraph{Static NeRFs.}

In this task, we experiment with three NeRF vaiants, including the original MLP-based NeRF~\cite{mildenhall2021nerf}, TensoRF~\cite{chen2022tensorf} and Instant-NGP~\cite{muller2022instant}. 
We show that NerfAcc works with both MLP-based and Voxel-based radiance fields, on both bounded (NeRF-Synthetic dataset~\cite{mildenhall2021nerf}, Tank\&Template dataset~\cite{Knapitsch2017}) and unbounded scenes (360 Dataset~\cite{barron2022mip360}). 
It is worth to note that with NerfAcc, one can train an Instant-NGP model with pure Python code and achieve slightly better performance than the official pure CUDA implementation, as shown in Table~\ref{tab:result_static}.

\paragraph{Dynamic NeRFs.}

In this task, we apply the NerfAcc toolbox to T-NeRF~\cite{pumarola2021d}, K-Planes~\cite{fridovich2023k} and TiNeuVox~\cite{fang2022fast}, covering both the synthetic (D-NeRF~\cite{pumarola2021d}) and ``in-the-wild'' captures\footnote{These datasets teleport cameras and do not represent real captures~\cite{gao2022monocular}.} (that accompany HyperNeRF~\cite{park2021hypernerf}). When applying the occupancy grid approach to accelerate those dynamic methods, instead of representing a static scene with it, we share the occupancy grid across all frames. In other words, instead of using it to indicate the opacity of an area at a single timestamp, We use it to indicate the \emph{maximum opacity at this area over all the timestamps}. This is not optimal but still makes the rendering very efficient, due to the fact there are limited movements in these datasets.

\paragraph{NeRFs for Camera Optimization.}

In this task, we employed the NerfAcc toolbox to BARF~\cite{lin2021barf} on the NeRF-Synthetic dataset with perturbed cameras. The goal is to jointly optimize the radiance field and camera extrinsic for multi-view images. We observed that the spatial skipping sampling provided by NerfAcc facilitated faster training and significantly improved both image quality and camera pose reconstruction. 
These improvements could be attributed to the sparsity enforced in our sampling procedure. This finding may provide interesting avenues for future research.

\begin{figure*}[!t]
\centering
\includegraphics[width=0.9\linewidth]{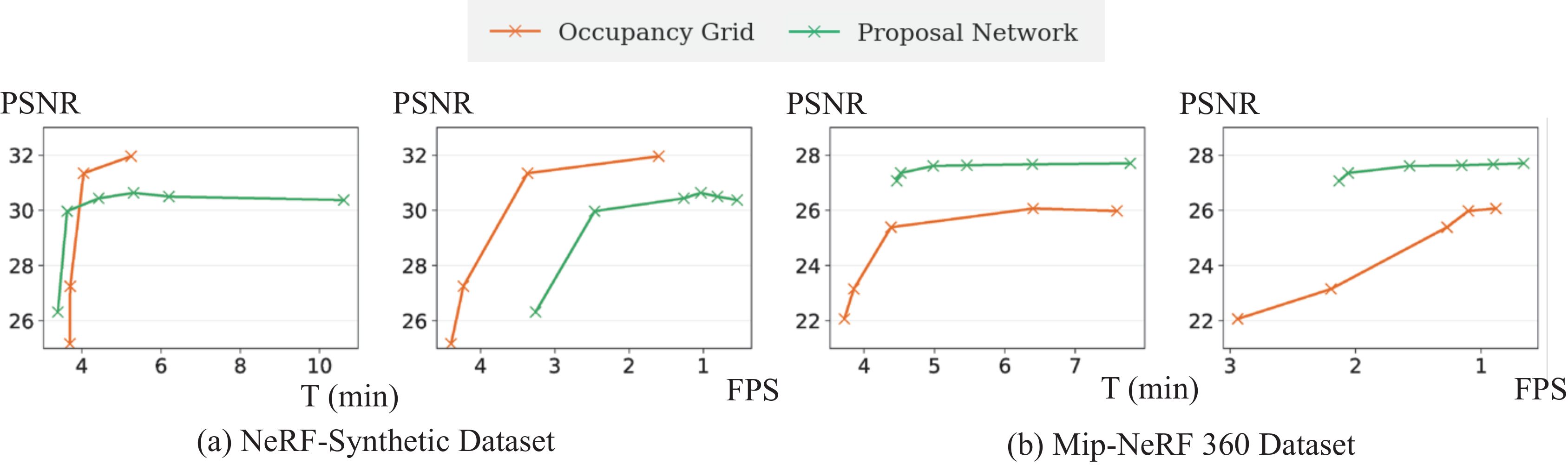}
\caption{\textbf{Comparison between Different Sampling Approaches {in terms of Training Time and Rendering FPS.}} We sweep the hyper-parameters space for each sampling approach, and find out that occupancy grid approach gives the best performance-speed trade-off on the NeRF-Synthetic dataset, while the proposal network approach performs the best on the Mip-NeRF 360 dataset. {Note T (min) denotes for \emph{training} time and FPS is for \emph{rendering} frames per second.} All experiments use the HashEncoding from Instant-NGP~\cite{muller2022instant} as the radiance field representation. Please see the supplementary materials for the hyperparameter space that we explored.
}
\vspace{-0.5em}
\label{fig:ray_marching}
\end{figure*}
\paragraph{Analysis of Different Sampling Approaches.} 
Results in Table~\ref{tab:result_static} show that the choice between occupancy grid and proposal network sampling can noticeably affect run-time and performance on different datasets. As each approach relies on a distinct set of hyperparameters, a systematic comparison between the two methods is crucial by sweeping the hyperparameter space. We varied the resolution and marching step size for occupancy grid and the number of samples and size of the proposal network for the proposal network approach. We plot pareto curves for each approach for both the NeRF-Synthetic and Mip-NeRF 360 datasets in Fig.\ref{fig:ray_marching}. This analysis indicates that occupancy grid sampling is suitable for the NeRF-Synthetic dataset, whereas the proposal network approach performs better on the 360 dataset. This is likely because the NeRF-Synthetic dataset contains more empty space that can be skipped effectively using the occupancy grid approach. However, in the case of real, unbounded data, the use of the occupancy grid approach is limited by the bounding box and the lack of empty space to skip, making the proposal network approach more effective. These experiments used the radiance field from Instant-NGP~\cite{muller2022instant}, with the same training recipes.

\subsection{Combined Sampling}
\label{ref:combined_smapling}
\begin{figure}[!t]
\centering
\includegraphics[width=0.9\linewidth]{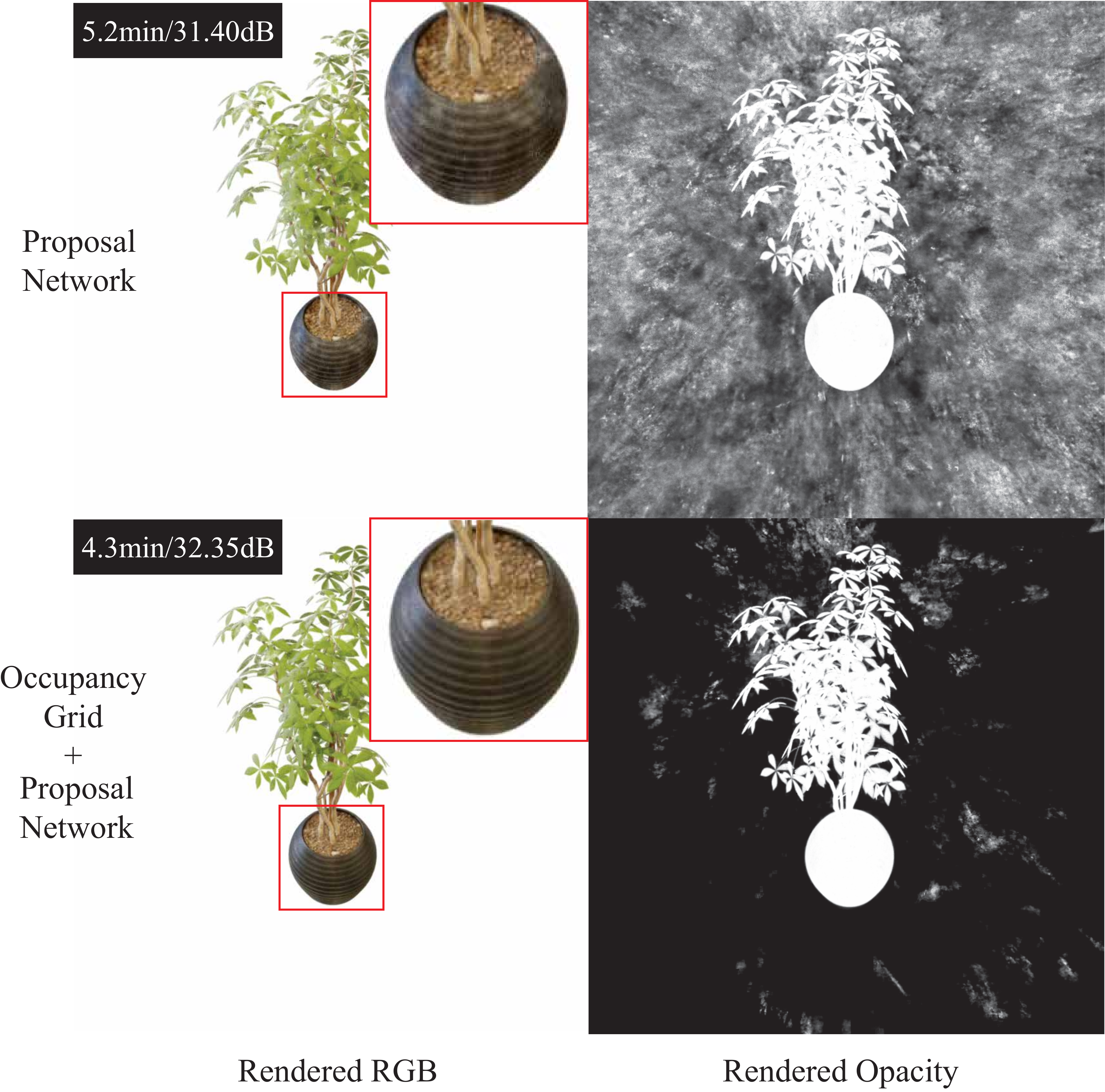}
\caption{\textbf{Results of Combined Sampling on NeRF-Synthetic Dataset.} Benefit from our unified concept of transmittance estimator, we find it's straight-forward to
combine the two distinct sampling approaches.
}
\label{fig:floaters}
\vspace{-0.5em}
\end{figure}

A benefit from the unified concept of transmittance estimator introduced in Section~\ref{sec:method}, is that it's straight-forward to combine the two distinct sampling approaches, as both of them essentially provide an estimation of the transmittance that can be used for importance sampling. For example, we find that simply stacking an occupancy grid on top of the proposal network, can significantly reduce the number of rays and shrink the near-far plane for the remaining rays on the NeRF-Synthetic dataset. This leads to improvements in quality, from $31.40\text{dB}$ to $32.35\text{dB}$, and a reduction in training time, from $5.2\text{min}$ to $4.3\text{min}$, compared to using only the proposal network for importance sampling. Figure~\ref{fig:floaters} shows an example with the \textsc{Ficus} scene, where the floaters are cleaned out with the combined sampling. This experiment is conducted using the HashEncoding from Instant-NGP~\cite{muller2022instant} as the radiance field representation.

\section{Conclusions}

In conclusion, this paper highlights the significant impact of advanced sampling approaches on improving the efficiency of Neural Radiance Fields (NeRF) optimization and rendering. We demonstrate that advanced sampling can significantly speed up the training of various recent NeRF papers, while maintaining high-quality results. The development of NerfAcc, a flexible Python toolbox, enables researchers to incorporate advanced sampling methods into NeRF-related methods easily. The exploration and comparison of advanced sampling methods are important steps towards developing more efficient and accessible NeRF-based methods. The presented results also demonstrate the potential for further research to improve the performance of NeRF and other related techniques through advanced sampling strategies.

\section*{Acknowledgement}
This project was supported in part by the Bakar Fellows Program and the BAIR/BDD sponsors.

{\small
\bibliographystyle{ieee_fullname}
\bibliography{egbib}
}

\newpage
\appendix
\section{Implementation Details for Case Studies}

This section describes the usage of NerfAcc to enhance multiple NeRFs, as discussed in Section 4.3 of the main paper. We provide detailed information on how to adapt NerfAcc to various NeRF models.

\subsection{Static NeRFs.}

\paragraph{Vanilla NeRF~\cite{mildenhall2021nerf}.} 
\looseness=-1 We trained an 8-layer MLP with the same structure as the vanilla NeRF paper. The paper employs the PDF-based efficient sampling approach with two MLPs for course-to-fine sampling (64 + 128). Although the PDF-based approach effectively concentrates samples around the surface, the MLP itself is computationally slow. To accelerate the process, we used the spatial skipping approach with Occupancy Grid provided by NerfAcc. Due to the considerable amount of empty space in NeRF-Synthetic dataset, we are able to increase the number of samples per ray to 1024 without any memory issue. As a result, we achieved a $20\times$ speedup and improved image quality by $+0.5\text{dB}$.

\paragraph{TensoRF~\cite{chen2022tensorf}.} 
\looseness=-1 Interestingly, incorporating NerfAcc into its own repository results in a $1.5\times$ speedup for TensoRF~\cite{chen2022tensorf}, a more recent voxel-based method, on both the NeRF-Synthetic and Tank-and-Temple datasets. We conjecture that this improvement is due to our update to their transmittance estimator. Although not discussed in its paper, TensoRF initially employs a skip-based transmittance estimator when sampling, similar to our occupancy grid. However, unlike our implementation, TensoRF's estimator does not maintain an accurate estimation of the main density field and only updates twice throughout the entire training process. Our approach, on the other hand, distills a running average of the density field every few training steps.

\paragraph{Instant-NGP~\cite{muller2022instant}.} 
\looseness=-1 Furthermore, we were able to reproduce Instant-NGP~\cite{muller2022instant} on the NeRF-Synthetic dataset with the same training speed and slightly better performance ($+0.2\text{dB}$) using the Occupancy Grid. We also achieved significantly better results on the Mip-NeRF 360 dataset using the proposal network approach. It is worth noting that the original Instant-NGP implementation is in pure CUDA with all operations fused into the CUDA kernels. With NerfAcc handling the underlying sampling logic, the entire training pipeline can be implemented in Python on top of PyTorch. We believe that our toolbox strikes a balance between high-performance computing and simplicity, benefiting rapid research development.

\subsection{Dynamic NeRFs.}

\paragraph{T-NeRF~\cite{pumarola2021d}} 
\looseness=-1 We trained a T-NeRF model described in the D-NeRF paper~\cite{pumarola2021d} on the D-NeRF dataset. By utilizing the occupancy grid in NerfAcc, we achieved a remarkable speedup of $20\times$ with a corresponding $+3.5\text{dB}$ improvement in performance. It is worth noting that we performed this experiment using our re-implementation of T-NeRF model and training recipe.

\paragraph{K-Planes~\cite{fridovich2023k}.}
\looseness=-1 For K-Planes, we achieve around 1.6$\times$ speedup on D-NeRF dataset by replacing its original proposal-based transmittance estimator with the occupancy grid. We believe that the sparsity in the D-NeRF dataset makes it more suitable for spatial skipping approaches. 

\paragraph{TiNeuVox~\cite{fang2022fast}.}
\looseness=-1 Our experiments with TiNeuVox on both datasets suggest a universal speedup by using our toolbox. Specifically, NerfAcc speeds up training by $2.8\times$ on D-NeRF and $1.7\times$ on HyperNeRF datasets, respectively. The original TiNeuVox implementation extends upon DVGO~\cite{sun2022direct} and directly marches rays within the main density grid, which can be wasteful since space skipping does not require precise geometry. Our occupancy grid overcomes this issue and has been shown to be more effective. Additionally, we found that a time-conditioned proposal network can achieve similar performance on the real bounded scenes in the HyperNeRF dataset, which aligns with what we observed in the Instant-NGP experiments with the proposal network.

\subsection{Camera Optimization.}
\paragraph{BARF~\cite{lin2021barf}.}
\looseness=-1 In the original implementation, BARF uniformly samples within the near-far range for each ray, analogous to the course level in the vanilla NeRF~\cite{mildenhall2021nerf}. We integrated the occupancy grid from our toolbox for spatial skipping, resulting in a 4$\times$ speedup during training. Additionally, this led to improved image quality ($+1.3\text{dB}$) and roughly $2\times$ lower camera registration error.

\section{Hyper-parameter Space}
\looseness=-1 Figure 5 in the main paper demonstrates the significant impact of hyper-parameters on the performance of each sampling approach. The occupancy grid approach relies on several primary hyper-parameters, such as the binary threshold $\tau$, the grid resolution $L^3$, and the marching step size $\Delta t$. Meanwhile, the proposal network approach involves hyper-parameters related to the setup of the proposal network(s) $F_\Theta$, as well as the number of samples $N$ to be taken along each ray. To ensure the robustness of our experiments, each data point in Figure 5 is trained with a random combination of hyper-parameters drawn from a reasonable range for each dataset.

\end{document}